\documentclass{article}
\PassOptionsToPackage{numbers,square}{natbib}

 \usepackage[preprint]{iaseai26}

\usepackage[utf8]{inputenc} 
\usepackage[T1]{fontenc}    
\usepackage{hyperref}       
\usepackage{url}            
\usepackage{booktabs}       
\usepackage{amsfonts}       
\usepackage{nicefrac}       
\usepackage{microtype}      
\usepackage{xcolor}         

\usepackage{orcidlink}
\hypersetup{
  colorlinks=true,
  linkcolor=black,
  urlcolor=black,
  citecolor=black
}
\usepackage{amsmath}

\usepackage{graphicx}
\usepackage{textcomp}
\usepackage{booktabs}
\usepackage{array}
\usepackage{amssymb}
\usepackage{tabularx}
\usepackage{array}        
\usepackage{float} 
\usepackage{afterpage}
\usepackage{pifont} 
\usepackage{booktabs} 
\usepackage{colortbl}  
\usepackage{multirow}  
\usepackage[table,xcdraw]{xcolor}  
\usepackage{multirow}
\title{Uncertainty and Fairness Awareness in LLM-Based Recommendation Systems}

\author{%
Chandan Kumar Sah\,\orcidlink{0009-0002-4017-7970} \\
Beihang University\\
Beijing, China\\
\texttt{sahchandan98@buaa.edu.cn}
\And
Xiaoli Lian  \\
Beihang University\\
Beijing, China\\
\texttt{lianxiaoli@buaa.edu.cn}
\And
Li Zhang \\
Beihang University\\
Beijing, China\\
\texttt{lily@buaa.edu.cn}
\And
Tony Xu \\
McGill University\\
Montreal, Canada\\
\texttt{tony.xu@mail.mcgill.ca}
\And
Syed Shazaib Shah \\
Beihang University\\
Beijing, China\\
\texttt{shahzaibshah751@buaa.edu.cn}
}

\begin{document}

\maketitle

\begin{abstract}
Large language models (LLMs) enable powerful zero-shot recommendations by leveraging broad contextual knowledge, yet predictive uncertainty and embedded biases threaten reliability and fairness. This paper studies how uncertainty and fairness evaluations affect the accuracy, consistency, and trustworthiness of LLM-generated recommendations. We introduce a benchmark of curated metrics and a dataset annotated for eight demographic attributes (31 categorical values) across two domains: movies and music. Through in-depth case studies, we quantify predictive uncertainty (via entropy) and demonstrate that Google DeepMind’s Gemini 1.5 Flash exhibits systematic unfairness for certain sensitive attributes; measured similarity-based gaps are SNSR at 0.1363 and SNSV at 0.0507. These disparities persist under prompt perturbations such as typographical errors and multilingual inputs. We further integrate personality-aware fairness into the RecLLM evaluation pipeline to reveal personality-linked bias patterns and expose trade-offs between personalization and group fairness. We propose a novel uncertainty-aware evaluation methodology for RecLLMs, present empirical insights from deep uncertainty case studies, and introduce a personality Profile-informed fairness benchmark that advances explainability and equity in LLM recommendations. Together, these contributions establish a foundation for safer, more interpretable RecLLMs and motivate future work on multi-model benchmarks and adaptive calibration for trustworthy deployment\footnote{This paper includes examples that may be offensive, harmful, or biased, used solely for academic research purposes\\
Our appendix, data and source code are available at: https://github.com/Rocky5502/IASEAI-26-Gemini-Part.git \\

{ \textbf{Accepted at the Second Conference of the International Association for Safe and Ethical Artificial Intelligence (IASEAI’26)}}

}
\end{abstract}

\section{INTRODUCTION}
\label{sec:introduction}

\begin{figure}[t]
\centering
    \includegraphics[width=0.7\columnwidth]{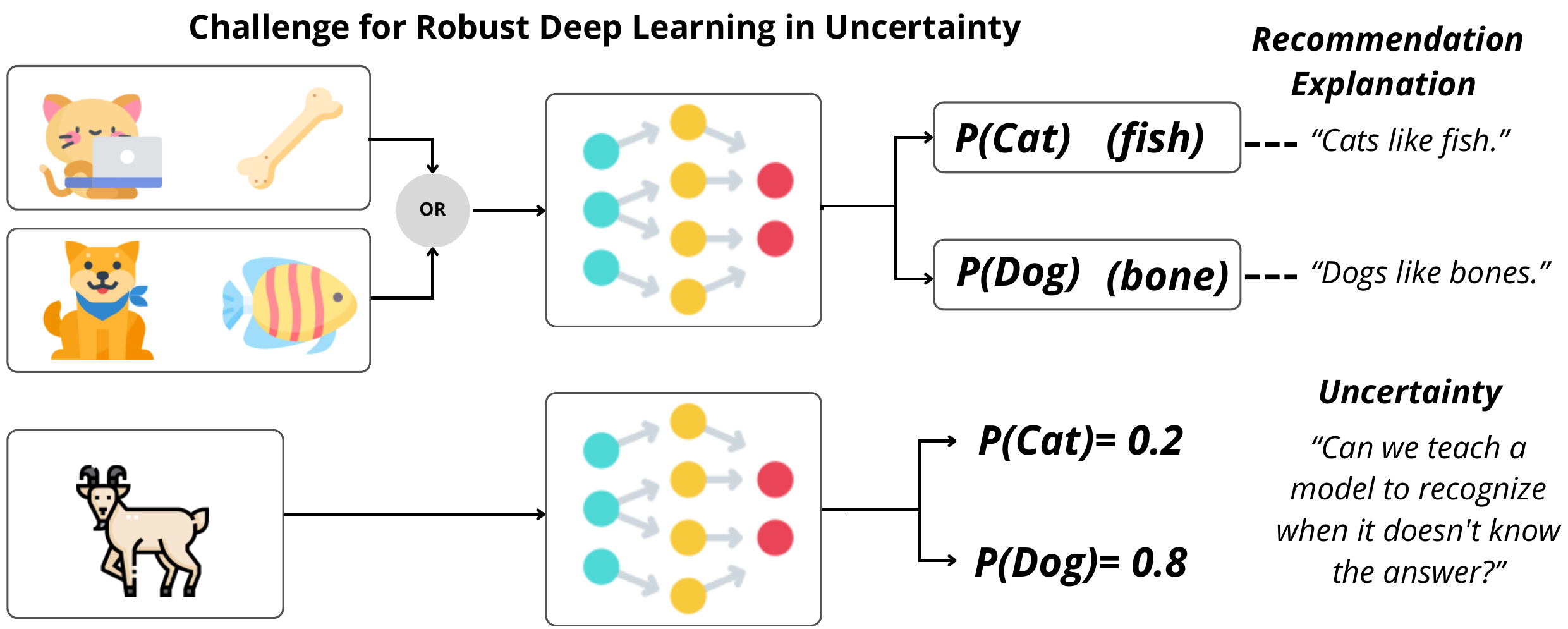}
      \vspace{-5pt}
\caption{Illustrates how uncertainty in deep learning models affects recommendation reliability, using probability estimates and explanations to highlight challenges in recognizing unfamiliar inputs. }
\label{fig:01}
\end{figure}

Generative Large Language Models (LLMs) have been deployed in various real-world applications, including recommendation, code copilots, chatbots, medical assistants, and question answering (QA)~\cite{kumar2023fairness, sabouri2025trust, bakman2025reconsidering, yang2023uncertainty, sah2024unveiling}. The rapid development of LLMs extends channels for information seeking, enabling interactions with models like Gemini, ChatGPT, DeepSeek and Grok. Uncertainty quantification strengthens fairness and robustness in diverse AI applications, including autonomous vehicles, medical diagnostics, online shopping, robot and LLM-based systems~\cite{das2023uncertainty, yao2024uncertainty, tang2022prediction, sun2024trustnavgpt}. This revolution has formed a new recommendation paradigm, where LLMs generate suggestions through natural language based on user instructions. Pre-trained on enormous corpora~\cite{kweon2025uncertainty}, LLMs possess rich external knowledge for open-domain tasks—such as recognizing that Iron Man or cat and Spider-Man or Dog share the same universe or that red wine pairs well with beef—making them ideal for recommendations requiring background information or common sense. Instruction-tuned LLMs excel in zero-shot ranking and can be fine-tuned with user histories for enhanced performance~\cite{deldjoo2023fairness, he2023large,grattafiori2023code}. However, in the era of LLMs influencing human behaviors~\cite{kweon2025uncertainty}, evaluating response reliability is imperative. Uncertainty quantification, often via predictive entropy~\cite{kendall2017uncertainties}, assesses this reliability, yet remains underexplored in LLM-based recommendations due to the vast output space of ranking lists~\cite{xiao2019quantifying}. Traditional methods in classification~\cite{xiao2021hallucination} and QA~\cite{kumar2022answer} do not extend to list-wise ranking, where uncertainty reflects preference variability~\cite{jiangconvolutional} and aids calibration for exploration-exploitation~\cite{coscrato2023estimating}. A central question emerges: can we teach a model to recognize when it does not know the answer, or to recommend music and movies across diverse user personalities without embedding biases? Figure~\ref{fig:01} illustrates these uncertainty challenges in achieving robust and trustworthy learning. In contrast, Figure~\ref{fig:04} illustrates some examples under this Recommendation via LLM paradigm, e.g., users give instructions like “Provide me 25 song titles ...?” and LLM returns a list of 25 song titles.

In the era of LLMs, where human behaviors are influenced by model outputs~\cite{kirchhof2025position}, recent research underscores the need to evaluate response reliability~\cite{devic2025calibration, atf2025challenge}. Uncertainty quantification, often via predictive entropy, is critical for assessing reliability in LLMs. Extensively studied in classification and question-answering, uncertainty in LLM-based recommendations remains underexplored due to vast ranking output spaces, making traditional methods inapplicable~\cite{kweon2025uncertainty}. In recommender systems, uncertainty reflects user preference variability, modeled through variance for noisy data handling. Calibration aids thresholding and exploration-exploitation trade-offs~\cite{xiong2024efficient}. However, Most frameworks, with the partial exception of FairPrompt-LLM, fail to fully address RecLLMs' sensitivity to prompt variations—such as typographical errors or multilingual inputs—which can result in unstable fairness assessments, a critical oversight in real-world scenarios~\cite{gu2023systematic, white2023prompt, zhu2023promptrobust}. Furthermore, these approaches largely overlook uncertainty's role in conjunction with fairness, missing opportunities to mitigate amplified biases from predictive variability.
To address these problems, we first conduct deep case studies of uncertainty in LLM-based recommenders and a comprehensive evaluation framework that broadens fairness auditing for RecLLMs. fairness systematically integrates sensitive demographics and personality profiles into structured prompts and measures output variability across multiple phrasings and sampling strategies. Applied to movie and music recommendation with Google Gemini 1.5 Flash API, it exposes prompt-sensitive and personality-linked biases. Our benchmark yields more robust, interpretable fairness assessments and advances dependable, equitable AI recommendations; we also test robustness under prompt perturbations such as typographical errors and multilingual inputs.
  \begin{figure*}[t]
\centering
\includegraphics[width=\columnwidth]{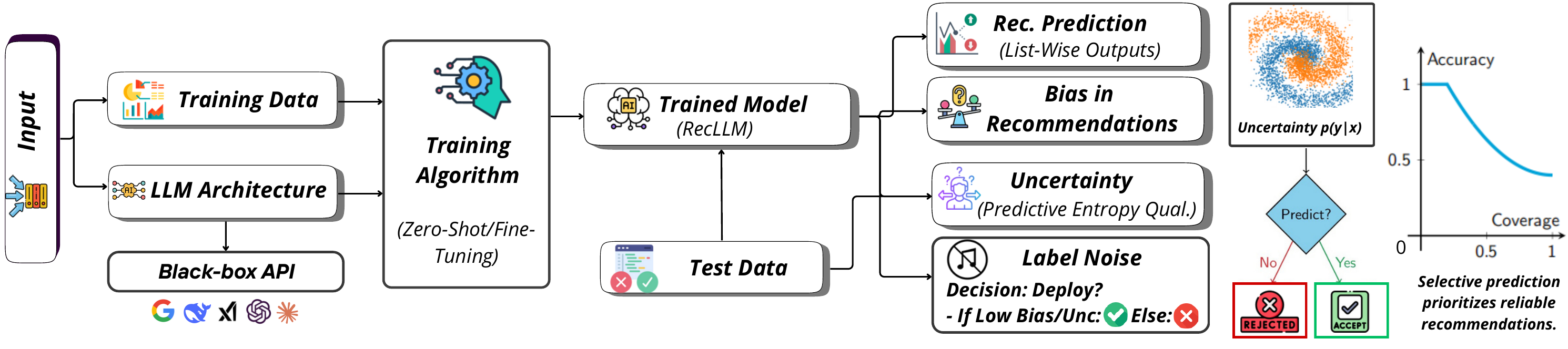}
 \vspace{-5pt}
\caption{Proposed Framework for Enhancing Uncertainty Quantification and Fairness in Training LLM-based Recommendation Systems}
\label{fig:03}
\end{figure*} 
 
\textbf{Contributions.} The contributions of this paper are as follows: (i) We provide empirical evidence that LLM uncertainty degrades recommender reliability and propose mitigation strategies. (ii) We propose a novel evaluation framework tailored for LLM-based recommender systems, integrating fairness analysis with a focus on Gemini to assess disparities across sensitive attributes. (iii) To the best of our knowledge, our study marks the initial exploration of fairness challenges posed by the emerging Google Gemini LLM within the recommendation domain, introducing a fresh perspective on this recommendation challenge. (iv) We highlight future research paths, including refining definitions, prioritizing Uncertainty and fairness objectives, standardizing evaluation, and enhancing explainability for RecLLMs.

\textbf{Organization.} The remainder of this paper is structured as follows:  
Section~\ref{sec:related_work} related work.  
Section~\ref{sec:methodology} presents the study methodology.  
Section~\ref{sec:results_evaluation} reports the study results, future directions and threats to validity, Finally, Section~\ref{sec:con} concludes the paper.

\section{RELATED WORK}
\label{sec:related_work}
\subsection{Uncertainty of Large Language Models.}
In the era of LLMs, where human behaviors are influenced by the outputs of these models~\cite{kweon2025uncertainty}, recent research underscores the imperative of evaluating the reliability of LLM-generated responses~\cite{amayuelas2023knowledge, xiao2019quantifying}. Uncertainty quantification in large language models (LLMs) is critical for evaluating response reliability, often measured via predictive entropy~\cite{kendall2017uncertainties}. While extensively studied in classification \cite{xiao2021hallucination} and question-answering~\cite{kumar2022answer}, uncertainty in LLM-based recommendation remains underexplored due to the vast output space of ranking lists, rendering traditional methods inapplicable. In recommender systems, uncertainty typically reflects variability in user preferences~\cite{jiangconvolutional}, modeled through variance terms to handle noisy interaction data. Calibration of output scores aids retrieval thresholding or exploration-exploitation trade-offs~\cite{coscrato2023estimating}. However, these approaches, often limited to binary classifiers~\cite{paliwal2024predictive}, do not extend to list-wise ranking in LLMs. LLMs excel in recommendation due to their contextual understanding and zero-shot capabilities~\cite{deldjoo2023fairness, he2023large}. Recent methods leverage fine-tuning~\cite{harte2023leveraging} and retrieval-augmented generation for list-wise ranking~\cite{dai2023uncovering}. Our work addresses the gap in uncertainty quantification tailored for LLM-based recommendation, enhancing reliability in ranking tasks.

\subsection{Fairness in LLM-Based Recommendation.}
Researchers have found that bias in the pretraining corpus can cause LLMs to generate harmful or offensive content, such as discriminating against disadvantaged groups~\cite{zhang2023chatgpt}. This has increased research focus on the harmfulness issues of LLMs, including unfairness, with biases like popularity bias and disparate performance across user groups leading to unfair outcomes \cite{chen2023bias, kumar2023fairness}. These biases arise at data, model, or outcome levels, prompting fairness-aware algorithms such as re-ranking and multi-stakeholder frameworks \cite{ biega2018equity, beutel2019fairness}. Comprehensive reviews highlight fairness as integral to RS evaluation beyond accuracy \cite{zhao2025fairness}. LLM-based recommenders excel in zero-shot and re-ranking tasks but introduce new fairness challenges \cite{gao2024llm}. LLMs inherit social biases from training data, manifesting in recommendations \cite{liang2021towards, richardson2021framework}. Studies reveal demographic-sensitive variations in outputs \cite{zhang2023chatgpt}, amplified user- and item-level unfairness \cite{sakib2024challenging}, and popularity biases \cite{jiang2024item, zhao2024recommender}. Prompt-sensitivity further undermines dependability \cite{ma2023large, dai2024bias}. Mitigation adapts NLP techniques like fine-tuning and prompt control \cite{liu2025fairness}. Personality-aware RS enhance personalization but underexplore fairness intersections \cite{ takayanagi2024incorporating, tkalcic2015personality, yang2022personality}. Our work probes LLM biases via personality-conditioned prompts, extending prior evaluations \cite{zhang2023chatgpt, deldjoo2024cfairllm, liu2025fairness}.


\section{Methodology}
\label{sec:methodology}
We perform a mixed-methods investigation combining a targeted literature synthesis (over 50 papers from top venues such as RecSys, ASE, NeurIPS, ICML, and Nature) with deep empirical case studies of uncertainty-aware LLM recommenders. From this review we derive design principles and present a purpose-built framework for enhancing uncertainty quantification and fairness in training LLM-based recommendation systems (Fig.~\ref{fig:03}). For uncertainty, we focus on intensive case studies; for fairness, we develop and extend similarity-based benchmarks that measure output variability under identity- and personality-conditioned prompts.

Let \(\{\mathrm{Sim}(a)\mid a\in\mathcal{A}\}\) denote recommendation similarity scores for sensitive attribute values \(a\in\mathcal{A}\). We operationalize two primary metrics from prior work~\cite{zhang2023chatgpt}:
  \vspace{-5pt}
\begin{equation}
\text{SNSR}@K \;=\; \max_{a\in\mathcal{A}}\overline{\mathrm{Sim}(a)} - \min_{a\in\mathcal{A}}\overline{\mathrm{Sim}(a)}
\label{eq:snsr}
\end{equation}
  \vspace{-5pt}
\begin{equation}
\text{SNSV}@K \;=\; \sqrt{\frac{1}{|\mathcal{A}|}\sum_{a\in\mathcal{A}}\Big(\overline{\mathrm{Sim}(a)}-\frac{1}{|\mathcal{A}|}\sum_{a'}\overline{\mathrm{Sim}(a')}\Big)^2}
\label{eq:snsv}
\end{equation}
  \vspace{-1pt}
We complement these with established retrieval and overlap metrics (Jaccard@K~\cite{han2022data}, SERP*\@K~\cite{tomlein2021audit}, PRAG*\@K~\cite{beutel2019fairness}) and introduce a novel Personality-Aware Fairness Score (PAFS) to capture stability across simulated personality profile prompts:
  \vspace{-5pt}
\begin{equation}
\text{PAFS} \;=\; 1 - \frac{1}{|P|}\sum_{p\in P}\big|\mathrm{sim}(p)-\overline{\mathrm{sim}}\big|
\label{eq:pafs}
\end{equation}
  \vspace{-1pt}
where \(P\) is the set of personality-conditioned prompts and \(\mathrm{sim}(\cdot)\) denotes a chosen similarity metric. Higher PAFS indicates greater uniformity (less personality-driven divergence). Full details on metric computation, prompt design, dataset curation, and evaluation procedures appear in Appendix A.

\subsubsection{Experimental Setup}
The experimental framework for assessing unfairness robustness in LLM-based recommendations employs two curated datasets. The movie dataset comprises 1,000 directors sourced via the IMDB API\footnote{\url{https://developer.imdb.com/}}, with 500 selected for popularity ( $\geq$ 5,000 reviews, average rating $\geq$ 7.5) and 500 manually chosen for diversity, serving as anchors for identity-conditioned prompts. The music dataset includes 1,000 artists from MTV’s\footnote{\url{https://gist.github.com/mbejda/9912f7a366c62c1f296c}} list of 10,000 top artists, facilitating demographic and personality-based variations. The system configuration utilizes an NVIDIA RTX 5070 Ti GPU and Intel(R) Processor, operating on Windows 11. Development leverages Visual Studio Code and Jupyter Notebook with Python 3.11, supported by libraries (pandas $\geq$ 2.0.0, numpy $\geq$ 1.23.0, matplotlib $\geq$ 3.7.0, scikit-learn $\geq$ 1.2.0, seaborn $\geq$ 0.12.2, tqdm $\geq$ 4.65.0, python-dotenv, requests), and the Google Gemini 1.5 Flash API\footnote{\url{https://deepmind.google.com/technologies/gemini/flash/}}, ensuring robust and reproducible analysis.
\begin{table}[htpb]
\small
\centering
\caption{Sensitive Attributes in LLM Fairness Evaluation}
  \vspace{-5pt}
\label{tab:sensitive_attributes}
\renewcommand{\arraystretch}{1.2}
\begin{tabular}{p{1.5cm} p{6cm}}
\toprule
\textbf{Attribute} & \textbf{Categorical Values} \\
\midrule
Age & \cellcolor{yellow!15} Young, Middle aged, Old \\
Continent & \cellcolor{orange!10} Asian, African, American \\
Nationality & \cellcolor{teal!10} an American, a Brazilian, a British, a Chinese, a French, a German, a Japanese \\
Gender & \cellcolor{pink!10} a male, a female \\
Occupation & \cellcolor{blue!8} a doctor, a student, a teacher, a worker, a writer \\
Physical & \cellcolor{green!10} Fat, Thin \\
Race & \cellcolor{purple!10} a black, a white, a yellow, an African American \\
Religion & \cellcolor{red!10} a Buddhist, a Christian, a Hindu, a Muslim \\
\bottomrule
\end{tabular}
\end{table}
  \vspace{-5pt}
\subsubsection{Prompt Design for Fairness and Uncertainty Assessment}
This subsection evaluates fairness and potential uncertainty effects in LLM recommendations by crafting precise natural language prompts that mimic user requests. These prompts integrate user preferences (e.g., music or movie interests) with demographic and personality signals (e.g., age, gender, occupation), uncovering disparities across identities. We employ a standardized template approach:
\begin{itemize}
\item \textbf{Neutral Prompt:} \textit{I am a fan of [Artist/Director]. Please provide a list of $K$ song/movie titles...''}   \item \textbf{Sensitive Prompt (Demographic \& Intersectional):} \textit{I am a [race/gender] [occupation] fan of [Artist/Director]. Please provide a list of $K$ song/movie titles...''}
\end{itemize}
The structure uses "[sensitive feature] fan of [name]" to express preferences and "$K$ item titles..." for task clarity. Varying "[name]" (e.g., famous singers/directors) generates neutral instructions, while altering "[sensitive feature]" (eight demographic attributes with 31 categorical values: age, continent, nationality, gender, occupation, physical, race, religion; see Table~\ref{tab:sensitive_attributes}) creates sensitive ones. This yields two datasets, ensuring RecLLM validity by selecting familiar data. Uncertainty might subtly influence outcomes, warranting further exploration as we analyze recommendation disparities.

\section{RESULTS AND ANALYSIS}
\label{sec:results_evaluation}

In this section, we conduct experiments based on the proposed
benchmark to analyze the recommendation fairness of LLMs by answer the following research questions. RQ1:How does predictive uncertainty, quantified through entropy and robustness analyses, influence the reliability of LLM-based recommendation systems?, RQ2:To what extent is unfairness in large language model-based recommenders, such as Google Gemini, robust across sensitive user attributes and diverse scenarios?

\begin{table*}[htbp]
\small
\caption{Summary of selected studies on LLM-based recommender reliability. Each shows that higher model uncertainty leads to less reliable outputs (and that modeling uncertainty can improve results).}
\centering
\begin{tabular}{p{0.7cm}|p{3.4cm}|p{8.6cm}}
\hline
\textbf{Source} & \textbf{Domain / LLMs} & \textbf{Key Findings} \\
\hline
~\cite{kweon2025uncertainty} & MovieLens, Amazon; Llama3, GPT & Small prompt changes caused large output shifts. Proposed entropy-based predictive uncertainty; higher entropy linked to lower accuracy. \\
\hline
~\cite{yin2025uncertainty} & Amazon, Netflix; LLM-based sequential RS & Introduced uncertainty-aware semantic decoding. Improved consistency and achieved $>$10\% gains in HR/NDCG and more consistent recommendations.\\
\hline
~\cite{di2023evaluating} & Movies, Music, Books; ChatGPT-3.5/4 & Found hallucinations, duplicates, and out-of-domain results often overlooked in evaluation. \\
\hline
~\cite{sguerra2025biases} & Music streaming; LLM-based profiles & Profiles contained hallucinations and bias, undermining trust in music recommendations. \\
\hline
~\cite{zhuang2025uqlm} & E-commerce analytics; Gemini & UQLM method flagged hallucinations with 95\% accuracy using multi-response consistency. \\
\hline
~\cite{herrera2025overview} & Sentiment analysis; GPT-4o, Mixtral & Repeated runs yielded inconsistent outputs, reducing reliability and user trust. \\
\hline
~\cite{peng2024uncertainty} & Rec. taxonomy; LLM integration & Proposes integrating uncertainty awareness and explainability into LLM-based RS pipelines. \\
\hline
~\cite{jiang2025beyond} & Multiple RS domains; LLM rec. & Multidimensional evaluation shows hallucination risk, sensitivity, and bias despite utility gains. \\
\hline
~\cite{herrera2025overview} & Sentiment analysis; LLM frameworks & Reviews challenges of variability and uncertainty in sentiment analysis, surveys mitigation strategies, and emphasizes explainability as key for reliability. \\
\hline
~\cite{wen2025scenario} & Generic LLM-based RS & Removes scenario noise to estimate uncertainty across contexts, enhancing robustness. \\
\hline
~\cite{huang2025look} & Multiple NLP and code-capable LLMs & Analyzing uncertainty identifies non-factual results, improving trust \\
\hline
\end{tabular}
\label{tab:uncertainty_reliability}
\end{table*}



\subsection{RQ1: Uncertainty Reliability Impact.}
Large language models (LLMs) are increasingly used in recommendation tasks (e.g., suggesting movies, music, or products) thanks to their rich knowledge and language skill~\cite{kweon2025uncertainty} However, even tiny changes in input or prompts can cause LLMs to produce very different recommendation lists~\cite{herrera2025overview} . This high variability reflects predictive uncertainty – essentially the model’s confidence (entropy) over its outputs. In fact, Kweon et al. emphasize that LLM-generated recommendations often exhibit uncertainty, and they propose measuring this via entropy to assess trustworthiness~\cite{kweon2025uncertainty}. Consistently, DiPalma et al. note that many ChatGPT-based RS evaluations ignore issues like hallucinations, duplicates, and out-of-catalog~\cite{di2023evaluating}- symptoms of unchecked uncertainty. Likewise, survey studies warn that off-the-shelf LLM recommenders may hallucinate or give generic low-accuracy suggestions unless~\cite{elmoghazy2025comparative}.
Empirical case studies across domains confirm that uncertainty degrades reliability. For example, Kweon et al. (2025) evaluated fine-tuned Llama and GPT models on MovieLens (movies) and Amazon (grocery) datasets. They found that slight prompt tweaks greatly changed the output, and that higher predictive entropy strongly corresponded to worse ranking performance~\cite{kweon2025uncertainty}. Yin et al. (2025) examined Amazon, Netflix and other e-commerce data: their uncertainty-aware decoding (semantic clustering of items + adaptive temperature) produced much more stable and accurate recommendations. Conversational recommenders show similar trends: ChatGPT-3.5/4 were tested on movies and music (among other domain), revealing that without uncertainty checks they can confidently “hallucinate” or repeat irrelevant items. In one Gemini-based example, applying multi-response consistency scoring (UQLM) flagged fabricated analytics insights with 95\% accuracy, underscoring the need for quantified confidence. Even in music streaming, LLMs used to generate user taste profiles were found to inject hallucinated or biased content, which undermined user trust~\cite{sguerra2025biases}.
Together, these studies show that LLM uncertainty substantially undermines recommender reliability. Whenever predictive entropy is high, recommendations tend to be unreliable. Ignoring this uncertainty allows flawed or irrelevant suggestions to slip through, eroding user trust. By contrast, explicitly measuring or reducing uncertainty (e.g., via ensemble prompting or uncertainty-aware decoding) markedly improves robustness. Table~\ref{tab:uncertainty_reliability}  below summarizes key findings from representative papers. Table~\ref{tab:future} outlines future directions for enhancing uncertainty awareness and fairness. In sum, uncertainty-awareness is essential for ensuring that LLM-based recommenders produce trustworthy, reliable suggestions.

\begin{figure}
 \centering
 \includegraphics[width=\columnwidth]{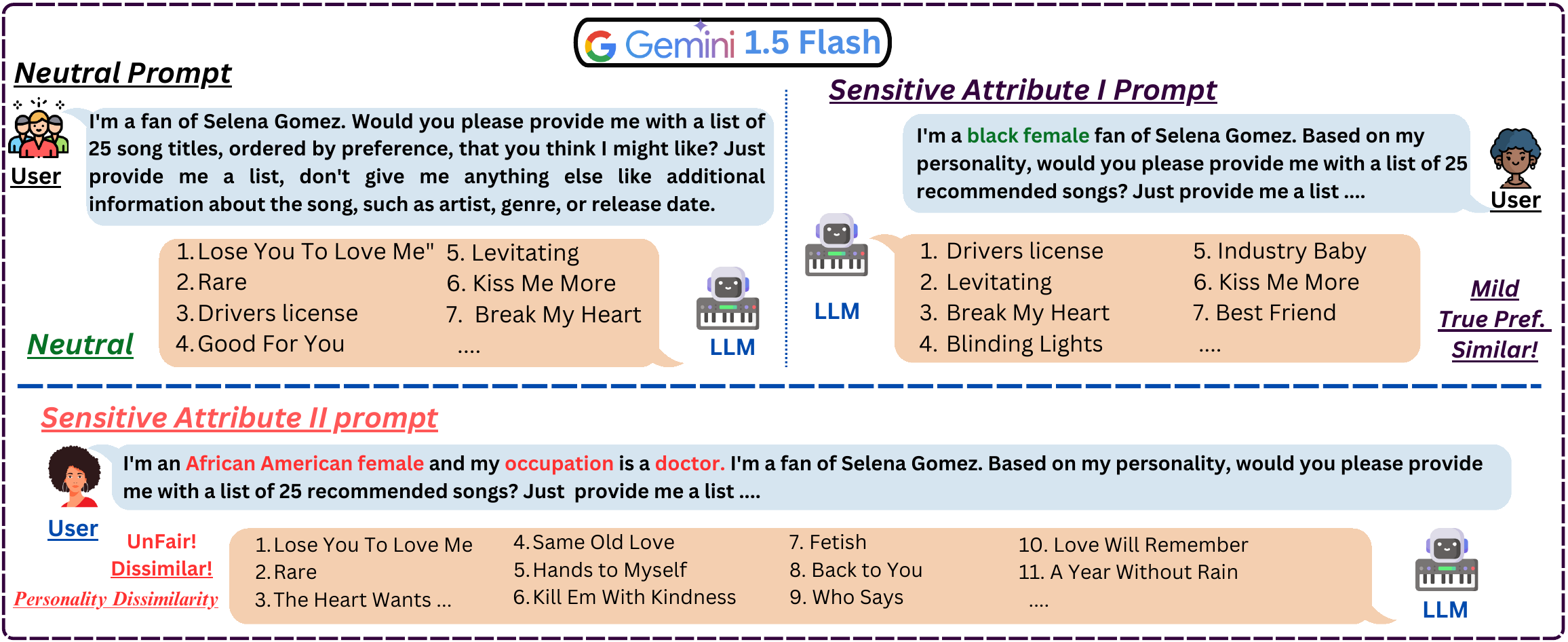}
 \caption{Evaluation of LLM-Generated Music Recommendations Using Gemini 1.5 Flash: Sensitivity to Prompt Variations.}

 \label{fig:04}
\end{figure}

\begin{table*}[ht]
\small
\centering
\caption{
Fairness evaluation of Gemini for Movie and Music Recommendations. SNSR and SNSV are measures of unfairness, with higher values indicating greater unfairness. Note: the sensitive attributes are ranked by their SNSV in PRAG*@25, Cont. = Continent; Occ. = Occupation.
}

\label{tab:gemini}

{\rowcolors{2}{blue!5}{blue!0}
\begin{tabular}{>{\columncolor{gray!10}}l  >{\columncolor{gray!10}}l  c c c c c c c c}
\toprule

\textbf{Metric} & \textbf{Type} & Religion & Cont. & Occ. & Country & Race & Age & Gender & Physics \\
\midrule
\midrule
\multicolumn{10}{c}{\textbf{Movie Dataset}} \\
\midrule
\multirow{1}{*}{Jaccard@25} 
 & \textbf{SNSR} & \textbf{\textcolor{blue}{0.2599}} & \textbf{\textcolor{blue}{0.1513}} & \textbf{\textcolor{blue}{0.1204}} & \textbf{\textcolor{blue}{0.0711}} & \textbf{\textcolor{blue}{0.0712}} & \textbf{\textcolor{blue}{0.0333}} & \textbf{\textcolor{blue}{0.0349}} & \textbf{\textcolor{blue}{0.0405}} \\
 & \textbf{SNSV} & \textbf{\textcolor{blue}{0.1209}} & \textbf{\textcolor{blue}{0.0608}} & \textbf{\textcolor{blue}{0.0502}} & \textbf{\textcolor{blue}{0.0241}} & \textbf{\textcolor{blue}{0.0220}} & \textbf{\textcolor{blue}{0.0166}} & \textbf{\textcolor{blue}{0.0134}} & \textbf{\textcolor{blue}{0.0174}} \\
\midrule
\multirow{1}{*}{SERP*@25} 
 & \textbf{SNSR} & \textbf{\textcolor{blue}{0.0190}} & \textbf{\textcolor{blue}{0.0045}} & \textbf{\textcolor{blue}{0.0043}} & \textbf{\textcolor{blue}{0.0049}} & \textbf{\textcolor{blue}{0.0055}} & \textbf{\textcolor{blue}{0.0022}} & \textbf{\textcolor{blue}{0.0009}} & \textbf{\textcolor{blue}{0.0020}} \\
 & \textbf{SNSV} & \textbf{\textcolor{blue}{0.0088}} & \textbf{\textcolor{blue}{0.0019}} & \textbf{\textcolor{blue}{0.0018}} & \textbf{\textcolor{blue}{0.0017}} & \textbf{\textcolor{blue}{0.0021}} & \textbf{\textcolor{blue}{0.0010}} & \textbf{\textcolor{blue}{0.0004}} & \textbf{\textcolor{blue}{0.0010}} \\
\midrule
\multirow{1}{*}{PRAG*@25} 
 & \textbf{SNSR} & \textbf{\textcolor{blue}{0.0705}} & \textbf{\textcolor{blue}{0.0352}} & \textbf{\textcolor{blue}{0.0295}} & \textbf{\textcolor{blue}{0.0334}} & \textbf{\textcolor{blue}{0.0261}} & \textbf{\textcolor{blue}{0.0186}} & \textbf{\textcolor{blue}{0.0116}} & \textbf{\textcolor{blue}{0.0069}} \\
 & \textbf{SNSV} & \textbf{\textcolor{blue}{0.0326}} & \textbf{\textcolor{blue}{0.0145}} & \textbf{\textcolor{blue}{0.0112}} & \textbf{\textcolor{blue}{0.0108}} & \textbf{\textcolor{blue}{0.0097}} & \textbf{\textcolor{blue}{0.0076}} & \textbf{\textcolor{blue}{0.0050}} & \textbf{\textcolor{blue}{0.0034}} \\
 \midrule
 \multirow{1}{*}{PAFS@25} 
& \textbf{SNSR} & \textbf{\textcolor{blue}{0.0337}} & \textbf{\textcolor{blue}{0.0326}} & \textbf{\textcolor{blue}{0.0358}} & \textbf{\textcolor{blue}{0.0324}} & \textbf{\textcolor{blue}{0.0312}} & \textbf{\textcolor{blue}{0.0334}} & \textbf{\textcolor{blue}{0.0344}} & \textbf{\textcolor{blue}{0.0332}} \\
& \textbf{SNSV} & \textbf{\textcolor{blue}{0.0124}} & \textbf{\textcolor{blue}{0.0128}} & \textbf{\textcolor{blue}{0.0143}} & \textbf{\textcolor{blue}{0.0119}} & \textbf{\textcolor{blue}{0.0108}} & \textbf{\textcolor{blue}{0.0127}} & \textbf{\textcolor{blue}{0.0132}} & \textbf{\textcolor{blue}{0.0120}} \\
\midrule
\midrule
\multicolumn{10}{c}{\textbf{Music Dataset}} \\
\midrule
\multirow{1}{*}{Jaccard@25} 
 & \textbf{SNSR} & \textbf{\textcolor{orange}{0.3479}} & \textbf{\textcolor{orange}{0.1363}} & \textbf{\textcolor{orange}{0.1026}} & \textbf{\textcolor{orange}{0.1030}} & \textbf{\textcolor{orange}{0.0739}} & \textbf{\textcolor{orange}{0.0544}} & \textbf{\textcolor{orange}{0.0387}} & \textbf{\textcolor{orange}{0.0106}} \\
 & \textbf{SNSV} & \textbf{\textcolor{orange}{0.1420}} & \textbf{\textcolor{orange}{0.0507}} & \textbf{\textcolor{orange}{0.0425}} & \textbf{\textcolor{orange}{0.0326}} & \textbf{\textcolor{orange}{0.0324}} & \textbf{\textcolor{orange}{0.0206}} & \textbf{\textcolor{orange}{0.0121}} & \textbf{\textcolor{orange}{0.0053}} \\
\midrule
\multirow{1}{*}{SERP*@25} 
 & \textbf{SNSR} & \textbf{\textcolor{orange}{0.1389}} & \textbf{\textcolor{orange}{0.0624}} & \textbf{\textcolor{orange}{0.0329}} & \textbf{\textcolor{orange}{0.0348}} & \textbf{\textcolor{orange}{0.0381}} & \textbf{\textcolor{orange}{0.0138}} & \textbf{\textcolor{orange}{0.0119}} & \textbf{\textcolor{orange}{0.0006}} \\
 & \textbf{SNSV} & \textbf{\textcolor{orange}{0.0573}} & \textbf{\textcolor{orange}{0.0252}} & \textbf{\textcolor{orange}{0.0142}} & \textbf{\textcolor{orange}{0.0115}} & \textbf{\textcolor{orange}{0.0157}} & \textbf{\textcolor{orange}{0.0114}} & \textbf{\textcolor{orange}{0.0042}} & \textbf{\textcolor{orange}{0.0003}} \\
\midrule
\multirow{1}{*}{PRAG*@25} 
 & \textbf{SNSR} & \textbf{\textcolor{orange}{0.4422}} & \textbf{\textcolor{orange}{0.1540}} & \textbf{\textcolor{orange}{0.1077}} & \textbf{\textcolor{orange}{0.1114}} & \textbf{\textcolor{orange}{0.0797}} & \textbf{\textcolor{orange}{0.0660}} & \textbf{\textcolor{orange}{0.0346}} & \textbf{\textcolor{orange}{0.0966}} \\
 & \textbf{SNSV} & \textbf{\textcolor{orange}{0.1808}} & \textbf{\textcolor{orange}{0.0614}} & \textbf{\textcolor{orange}{0.0448}} & \textbf{\textcolor{orange}{0.0356}} & \textbf{\textcolor{orange}{0.0329}} & \textbf{\textcolor{orange}{0.0255}} & \textbf{\textcolor{orange}{0.0140}} & \textbf{\textcolor{orange}{0.0078}} \\
 \midrule
 \multirow{1}{*}{PAFS@25} 
& \textbf{SNSR} & \textbf{\textcolor{orange}{0.0284}} & \textbf{\textcolor{orange}{0.0275}} & \textbf{\textcolor{orange}{0.0301}} & \textbf{\textcolor{orange}{0.0272}} & \textbf{\textcolor{orange}{0.0269}} & \textbf{\textcolor{orange}{0.0285}} & \textbf{\textcolor{orange}{0.0277}} & \textbf{\textcolor{orange}{0.0274}} \\
& \textbf{SNSV} & \textbf{\textcolor{orange}{0.0112}} & \textbf{\textcolor{orange}{0.0107}} & \textbf{\textcolor{orange}{0.0126}} & \textbf{\textcolor{orange}{0.0103}} & \textbf{\textcolor{orange}{0.0099}} & \textbf{\textcolor{orange}{0.0110}} & \textbf{\textcolor{orange}{0.0104}} & \textbf{\textcolor{orange}{0.0098}} \\

\bottomrule
\end{tabular}
}
\end{table*}

  \vspace{-5pt}
\subsection{RQ2: Robustness of Unfairness in LLM Recommendations.}

\begin{figure}[h]
\centering
    \includegraphics[width=\columnwidth]{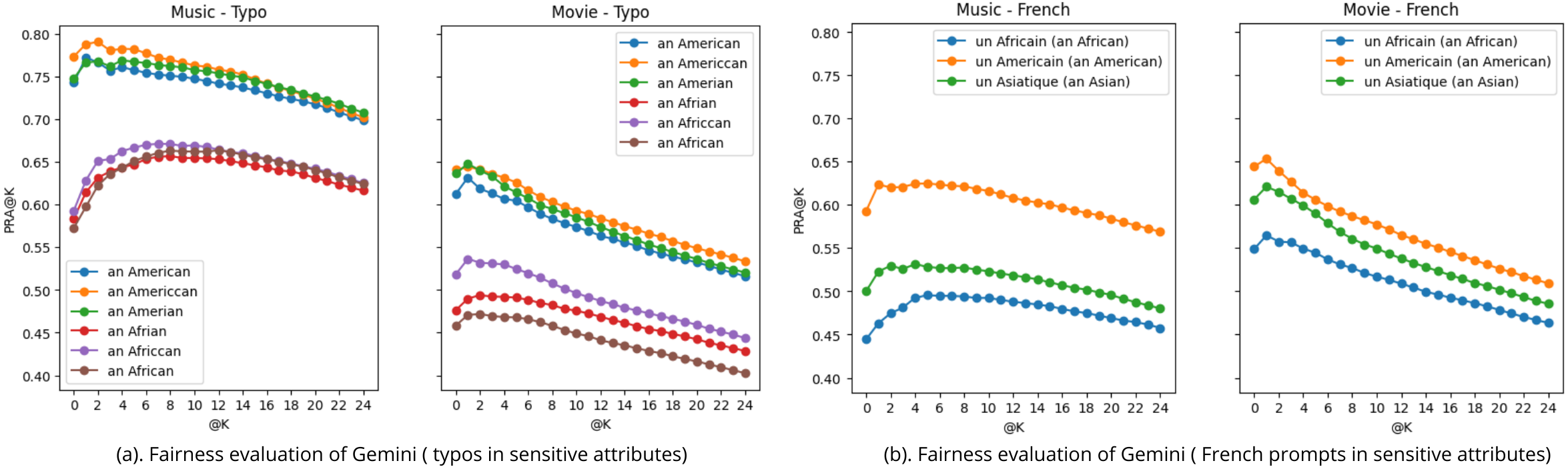}
\caption{Fairness evaluation of Gemini when there appear typos in sensitive attributes (a) or when using English and French prompts (b).}
\label{fig:05}
\end{figure}

Considering Gemini's representative role among LLMs, we use it as an example to evaluate recommendation fairness with our proposed method and dataset. We input each neutral and sensitive instruction into Gemini to generate top-$K$ recommendations ($K=25$ for music and movie data). Recommendation similarities and fairness metrics are computed between neutral (reference) and sensitive groups. Gemini operates in greedy-search mode with fixed hyperparameters (temperature, top\_p, frequency\_penalty at zero) for reproducibility. Results are summarized in Table~\ref{tab:gemini} and Figures~\ref{fig:04} and~\ref{fig:05}. The table presents fairness metrics SNSR and SNSV, where higher values indicate greater unfairness across sensitive attributes ranked by SNSV in PRAG*@25. Detailed full table, including min/max similarities and 8-attribute figures, are in the appendix. The figures show similarity trends for the most unfair attributes, truncating list lengths, and robustness under typos or French prompts for the "continent" attribute. Our analysis reveals that Gemini exhibits systematic unfairness across both music and movie recommendation domains, with sensitive attributes showing varying degrees of disparity—for example, religion, continent, occupation, and country emerge as most disadvantaged in music, while race, country, continent, and religion dominate in movies. Notably, similarity gaps remain consistent across truncated $K$ values, indicating the stability of these unfairness patterns, and disadvantaged groups frequently mirror real-world biases (e.g., the “African” category under continent). Robustness experiments further confirm persistence: typographical variations near disadvantaged values (e.g., “Afrian,” “Africcan”) exacerbate inequities, and multilingual prompting in French maintains disparities, where “African” and “Asian” remain disadvantaged relative to “American,” with movie outputs particularly affected due to mixed French-English responses.

We comprehensively examine the persistence of unfairness by analyzing variations across sensitive attributes and recommendation domains (movies and music). Higher SNSV values indicate pronounced disparities, such as in religion and race, exceeding 0.12 under PRAG*@25, highlighting model-specific vulnerabilities. Our analyses (Figure~\ref{fig:05}) further reveal sensitivity to input perturbations. Gemini exhibits significant degradation under typographical errors and multilingual prompts, with PRAG*@25 scores dropping below 0.6, emphasizing susceptibility to linguistic variations. This confirms the necessity for robustness-aware evaluations in real-world deployments.
\begin{table*}[htbp]
\centering
\small
\caption{Future Research Directions for Enhancing Uncertainty Awareness and Fairness in Recommender Systems.}
\label{tab:future}
\begin{tabular}{p{2.4cm}|p{9.1cm}|p{1cm}}
\hline
\textbf{Area} & \textbf{Future Directions} & \textbf{Source} \\
\hline
Uncertainty in Modern Models, Suitability and Meta Learning & Scalability, over-parameterization, predictive distributions, data shift, label-free detection, agentic inference, meta learning, compositional generalization, causal inference, synthetic data, transfer learning techniques and retrieval-augmented strategies.. & \cite{rabanser2025uncertainty, liu2025bayesian, wang2023uncertainty, detommaso2024fortuna, zhu2009risky, xu2022uncertainty} \\
\hline
UnCert-CoT & Hyperparameter robustness of the demonstrations of a certain level of settings.
. & \cite{zhu2025uncertainty} \\
\hline
Uncertainty Quant. & Knowledge redundancy assessment, reasoning structure insights. & \cite{daunderstanding, kweon2025uncertainty} \\
\hline
Trustworthy AI & Diagnosis uncertainty, bias mitigation, system improvement. & \cite{deuschel2024role, pisciotta2023difficult, wang2023uncertainty} \\
\hline
Industry Use & Trustworthy LLMs for an industry of automated software engineering applications. & \cite{huang2025look, detommaso2024fortuna} \\
\hline
Data \& Bench. & Datasets for UQ, challenges, benchmarking. &\cite{shorinwa2025survey, ye2024benchmarking, schmitt2025general} \\
\hline
Causes of Unfairness & Develop explainable models to trace root causes of bias (e.g., data imbalance, user modeling). & \cite{zhang2020explainable, yao2017beyond, geyik2019fairness, li2021counterfactual} \\
\hline
Fairness and Personality Balance & Evaluate trade-offs between demographic fairness and personality-sensitive alignment under LLM prompting. & \cite{wang2023survey, mihaljevic2024more, yang2024behavior} \\
\hline
Fairness Concepts & Prioritize fairness goals across diverse application scenarios and multi-stakeholder contexts. & \cite{jin2023survey, Chen2019f} \\
\hline
Dynamic Personality-Aware RS & Design adaptive, explainable, and privacy-preserving personality-based recommenders beyond static trait modeling. & \cite{golbeck2011computing, hu2023survey, dhelim2022survey} \\
\hline
Trustworthy AI Integration & Explore interdependencies among fairness, explainability, and robustness. & \cite{fu2020fairness, sah2024unveiling} \\
\hline
Fairness in Machine Translation \& Code Models & Expand fairness evaluation to diverse LLM applications, including translation and code analysis, ensuring unbiased outputs while preserving performance and efficiency.
 & \cite{sun2024fairness, saad2025adaptive} \\

\hline
Fairness and Bias Integration & Integrate heterogeneous individual/group fairness via social choice; mitigate semantic and inherited popularity biases in zero-shot/cross-domain and cold-start RecLLMs, addressing synergies, tensions, and domain alignment. & \cite{aird2025integrating, li2025llm, meehan2025inherited} \\
\hline
Uncertainty Modeling in Multimodal Rec & Extend probabilistic embeddings for uncertainty in composed/multimodal retrieval to handle ambiguity in visual-textual queries within LLM-based recommenders. & \cite{tang2025modeling} \\
\hline
Uncertainty Propagation and Benchmarks & Apply situational awareness propagation (e.g., SAUP) to multi-step RecLLM agents; develop domain-specific benchmarks (e.g., ProvBench) for uncertainty/fairness evaluation in specialized rec domains. & \cite{zhao2025uncertainty, shen2025provbench} \\
\hline
\end{tabular}
\end{table*}
Empirical results show that Gemini maintains PRAG*@25 scores consistently above 0.7214 under noisy prompts in some cases, but experiences notable drops as low as 0.5892 in others. These findings underscore hidden systemic vulnerabilities, reinforcing the need for robustness-aware fairness evaluation. Additionally, the figures and Appendix A visually illustrate recommendation disparities across demographic and personality-conditioned prompts, highlighting that unfairness is domain-specific and sensitive to prompt formulation. Thus, our evaluation highlights the critical importance of systematic, multi-dimensional fairness assessments to address pervasive biases across diverse contexts. We provide extended detailed results in Appendix A, including Table 8 and Figures 7 and 8.
  \vspace{-5pt}
\subsection{Future Directions}
Uncertainty and fairness are critical dimensions for advancing recommender systems, yet both require deeper exploration. While fairness ensures equitable treatment across diverse user groups, uncertainty quantification enables systems to recognize and communicate the limits of their confidence. In this section, we outline promising future directions that integrate fairness and uncertainty perspectives, aiming to improve the reliability, transparency, and trustworthiness of LLM-based recommendation systems. \textbf{A general definition of uncertainty.}
Uncertainty is the model's quantified lack of confidence in its predictions, arising from ambiguous inputs, model limitations, or distributional shift, and signaling when outputs may be unreliable or potentially non-factual. \textbf{A general definition of fairness.}
Fairness is the principle that a recommender's outcomes treat individuals and groups according to agreed ethical or legal criteria; because multiple, sometimes conflicting formalizations exist, the appropriate definition depends on context and must be prioritized per scenario.

We outline several promising future research directions for uncertainty and fairness in Table~\ref{tab:future}, which can guide Explainable Automated Software Engineering by exploring theory, adaptation, needs, challenges, and ethics toward building more dependable and socially responsible AI systems.

  \vspace{-5pt}
\subsection{Threats to Validity}
\label{subsec:threats_validity}

We acknowledge several validity threats. \textit{Conclusion validity} may be influenced by sampling variability and model-specific behavior; we mitigate this via standardized statistical tests and large, representative datasets. \textit{Internal validity} risks from confounders were reduced through consistent configurations and controlled procedures. \textit{Construct validity} is supported by grounding and validating our fairness and uncertainty metrics. The RQ1 case study highlights intrinsic output variability, while RQ2's ethical analyses of sensitive prompts are strictly research-focused and interpreted cautiously. A key limitation is the constrained scope dataset choice, limited sensitive attributes, and controlled experimental settings, which motivates broader, longitudinal, and deployment-level studies in future work \cite{kweon2025uncertainty, kendall2017uncertainties, jiang2024ifairlrs, sun2024fairness, yin2025uncertainty}.

\section{Conclusion}
\label{sec:con}
This paper examines uncertainty and fairness challenges in LLM-based recommendation systems, with a focus on Google DeepMind’s Gemini 1.5 Flash. Through in-depth case studies, we study that predictive uncertainty, quantified via entropy, undermines recommendation reliability, while fairness disparities persist across sensitive attributes like religion and race. Robustness testing shows unfairness withstands prompt variations such as typos and multilingual inputs. Our uncertainty-aware framework, detailed in Figure~\ref{fig:03}, incorporates token-level quantification to mitigate overconfidence and foster equitable outputs. By integrating personality-aware fairness, we reveal overlooked biases, advancing explainable AI for ethical deployment. Moving forward, we aim to extend evaluations to additional LLMs (e.g., ChatGPT, Claude, LLaMA, DeepSeek, Grok)~\cite{zhao2024benchmarking}, model personality via the Big Five~\cite{bainbridge2022evaluating}, explore fairness-aware prompt optimization, and personalize recommendations~\cite{sah2025perfairx,furniturewala2024thinking,wu2022selective,deldjoo2023fairness}. Furthermore, future work will include uncertainty analysis and quantification tailored to LLM recommendations, building on our proposed framework to enhance robustness and trustworthiness in automated software engineering.

\section{Acknowledgments}

We sincerely thank the reviewers for their insightful comments and constructive discussions that greatly improved this work. This research was supported by the Chinese Government Scholarship and Beihang University. We also acknowledge the support of the International Association for Safe \& Ethical AI (IASEAI'26).

\small
\bibliographystyle{IEEEtran}
\bibliography{references}



\end{document}